\documentclass[10pt,twocolumn,letterpaper]{article}
\usepackage{iccv}
\usepackage{times}
\usepackage{epsfig}
\usepackage{graphicx}
\usepackage{amsmath}
\usepackage{amssymb}

\usepackage{makecell}
\usepackage{multirow}
\usepackage[colorlinks]{hyperref}

\usepackage{epsfig}
\usepackage{graphicx}
\usepackage{amsmath,amssymb} 
\usepackage{color}
\usepackage{array}
\usepackage{bbm}
\usepackage{subcaption}
\usepackage{algorithm}
\usepackage{algpseudocode}
\usepackage{enumitem}
\usepackage{graphicx}
\usepackage{amssymb}
\usepackage{pifont}
\usepackage{amsfonts}
\usepackage{booktabs}
\usepackage{siunitx}
\usepackage{graphicx} 
\usepackage{epsfig}
\usepackage{float}
\usepackage{lipsum}
\usepackage{diagbox}
\usepackage{subcaption}
\usepackage{colortbl}
\usepackage[utf8]{inputenc}
\usepackage[english]{babel}
\usepackage[dvipsnames]{xcolor}
\usepackage{array,multirow,graphicx}

\usepackage[T1]{fontenc}
\usepackage{authblk}

\usepackage{color}

\iccvfinalcopy 

\def\iccvPaperID{****} 

\ificcvfinal\fi

\begin{document}


\title{Synthetic Data Are as Good as the Real for Association Knowledge Learning in Multi-object Tracking}

\author[1]{Yuchi Liu}
\author[2]{Zhongdao Wang}
\author[2]{Xiangxin Zhou}
\author[1]{Liang Zheng}

\affil[1]{Australian National University}
\affil[2]{Tsinghua University}

\maketitle
\ificcvfinal\thispagestyle{empty}\fi

\begin{abstract}
Association, aiming to link bounding boxes of the same identity in a video sequence, is a central component in multi-object tracking (MOT). 
To train association modules, e.g., parametric networks, real video data are usually used. However, annotating person tracks in consecutive video frames is expensive, and such real data, due to its inflexibility, offer us limited opportunities to evaluate the system performance w.r.t changing tracking scenarios.
In this paper, we study whether 3D synthetic data can replace real-world videos for association training. 
Specifically, we introduce a large-scale synthetic data engine named MOTX, where the motion characteristics of cameras and objects are manually configured to be similar to those in real-world datasets. 
We show that compared with real data, association knowledge obtained from synthetic data can achieve very similar performance on real-world test sets without domain adaption techniques. 
Our intriguing observation is credited to two factors. First and foremost, 3D engines can well simulate motion factors such as camera movement, camera view and object movement, so that the simulated videos can provide association modules with effective motion features. Second, experimental results show that the appearance domain gap hardly harms the learning of association knowledge.
In addition, the strong customization ability of MOTX allows us to quantitatively assess the impact of motion factors on MOT, which brings new insights to the community~\footnote{This work is available at \href{https://github.com/liuyvchi/MOTX}{https://github.com/liuyvchi/MOTX}.}. 

\end{abstract}

\section{Introduction}

Multi-object tracking (MOT) is a compound system composed of several functional components, \eg, detection, visual representations, and association. Association is at the final stage in the MOT pipeline and is usually viewed as the core problem, aiming to connect bounding boxes with existing tracklets~\cite{xu2020train, braso2020learning}. The association module makes inference according to appearance features (\eg re-identification features), motion features (\eg location and size of bounding boxes), or both of them.

In the community, what many solutions to the association have in common is that they are trained with real-world video data~\cite{leal2015motchallenge, milan2016mot16}. There are several potential problems with this practice. First, annotating trajectories in video frames requires expensive labor costs. 
This potentially limits the scale of MOT training data.  
Second, privacy and ethics issues constrain the usage of real-world data in human-centered tasks, \emph{e.g.}, multiple pedestrian tracking. 




\begin{figure}[t]
	\centering  
	\includegraphics[width=\linewidth]{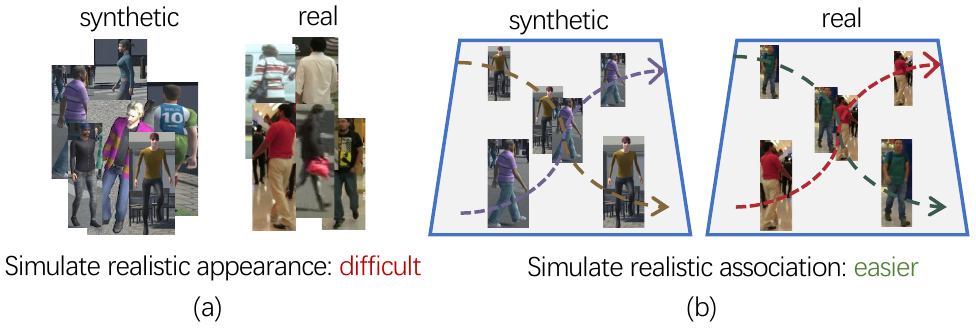}
	\caption{
	(a) Simulated appearance \vs (b) simulated association scenarios.~Simulated appearance usually has an image-style discrepancy with real-world appearance. For many appearance-centered tasks such as re-identification, such appearance domain gap compromises models that are trained on synthetic data and tested on real data. 
	In comparison, we show that synthetic data are as effective as real data in training association models. It suggests that association scenarios (\textit{e.g.,} trajectories, and occlusions) have a small domain gap between the synthetic and the real.
	}
 	\vspace{-1mm}
	\label{fig:simulation_difficulty}
\end{figure}

In this paper, we investigate how to use synthetic data in MOT, so as to avoid the concerns listed above. We build a 3D simulation engine, \textbf{MOTX}, for generating videos with multiple targets, rich annotations, and controllable visual factors. Such data offer an inexpensive way to acquire large-scale data with accurate labels. With MOTX, we aim to answer two interesting questions.


First, \emph{does association knowledge learned from synthetic data work in real-world videos?} 
A common weakness of synthetic data consists in its distribution difference with real-world data, especially regarding the image-style. In ``Appearance-centered'' tasks (\eg, re-identification, segmentation), to avoid failure in real-world test environments, models trained on synthetic data require additional training techniques, such as fine-tuning or domain adaptation on the real data~\cite{yao2019simulating, xue2020learning, dou2021versatilegait, bak2018domain}. 
However, association learning is different from appearance learning regarding data requirement. According to existing works~\cite{braso2020learning, xu2020train, li2020graph}, motion cues play an essential role for association. 
While appearance realistic images are hard to simulate by the engine, it may be less difficult for motion cues, such as occlusion. Some sample results of appearance simulation and association scenario simulation are shown in Fig.~\ref{fig:simulation_difficulty}. 

Second, \emph{how do motion factors affect association knowledge learning?}
Existing datasets are mostly from the real world, such as MOT15. While these data benefit model training, that they are fixed offers us limited opportunities to understand how the system reacts to changing visual factors. For example, how does pedestrian density in the training set affect model accuracy? Can a model trained with static cameras be well deployed under moving-camera systems? In this paper, taking advantage of the strong customization ability of MOTX, we will make some initial investigations on these interesting directions. 

In answering these two questions, this paper makes a two-fold contribution. First and most importantly, we show that on several state-of-the-art association networks, association knowledge learned from synthetic data can be well adapted to real-world scenarios without performance drop. Specifically, we synthesize datasets using MOTX by manually setting key parameters (\eg, camera view) to be close to real-world training sets.\footnote{Our manual parameter tuning is very efficient: a rough estimation of the motion parameters will do.} Then, when the recent association networks are trained on such synthetic videos, they achieve similar or sometimes even better tracking accuracy compared with real data training. Our ablation studies on appearance features and motion features suggest: 1) The appearance-discrepancy between synthetic data and real-world data can hardly harm the association knowledge learning. 2) 3D engines can well simulate the motion cues in association scenarios. The above findings can be the reason for the competitiveness of synthetic data, and implies that MOT benefits more from using synthetic data than ``Appearance-centered'' tasks.

Second, we perform empirical studies on how object-related and camera-related factors affect learning of association knowledge. Specifically, we investigate two groups of factors: 1) Pedestrian-related factors, such as density and moving speed; 2) and camera-related factors, including the camera view and camera moving state. In detail, with the proposed MOTX engine\footnote{This engine will be publicly released soon.}, motion factors are abstracted with system parameters, so we can readily simulate different scenarios by simply changing these parameters, \eg, set the object velocity to $1$m/s. 
Our results shed light on the relationship between factors in training and testing data and MOT system performance. 

\begin{figure*}[t]

	\centering  
	\includegraphics[width=\linewidth]{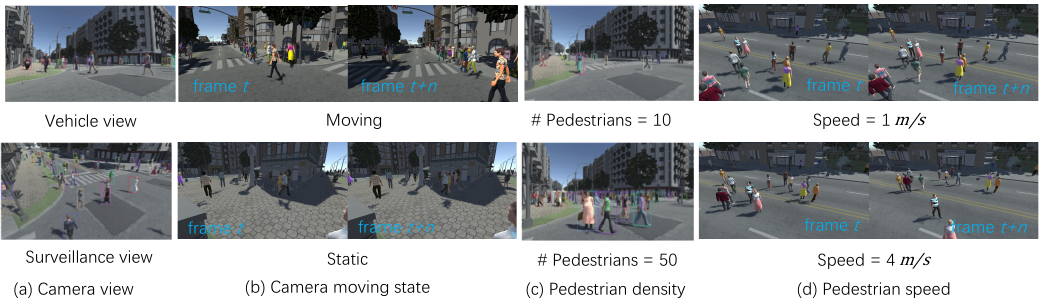}
	\caption{MOTX provides an inexpensive and accurate way to generate videos and their labels for association training. \emph{Controllable} factors include \textbf{(a)} Camera view, \textbf{(b)} Camera moving state, \textbf{(c)} Pedestrian density, and \textbf{(d)} Pedestrian speed.}
	\label{fig:motx_data}
\end{figure*}

\section{Related Work}
\textbf{Association methods in MOT.} There are mainly two types of association: human-designed policies and parametric association modules. The former is usually seen in MOT works focusing on improving detection and appearance embedding~\cite{wojke2017simple, zhu2018online, zhou2018online, li2020graph, zhan2020simple, wang2019towards}. They compute similarities between bounding boxes and objects according to predefined metrics. The most commonly used metrics are IoU score and Cosine similarity score between deep Re-ID features. 
Then a bipartite matching algorithm (\textit{e.g.,} the Hungarian algorithm~\cite{kuhn1955hungarian}) associates bounding boxes with objects. Kalman Filterd~\cite{welch1995introduction} can also predict motion and smooth trajectories. 


The latter uses neural networks to formulate the association stage.  
For example, DeepMOT~\cite{xu2020train} proposes a LSTM method to approximate the Hungarian matching algorithm~\cite{kuhn1955hungarian}. MPNTracker~\cite{braso2020learning} formulates sequences as graphs and designs a differentiable message passing network to predict the score for each box link between frames. Li \emph{et al.}~\cite{li2020graph} and Papakis \emph{et al.}~\cite{papakis2020gcnnmatch} use a graph neural network to model appearance and motion (geometric) features and produce the similarities between tracklets and detections. These parametric association modules are trained based on appearance features and motion features. 
In this paper, we observe that parametric association modules trained with synthetic videos can be successfully deployed in real-world test sets without domain adaptation.

\textbf{Learning from synthetic data for real-world applications.} Synthetic datasets have been used in image classification~\cite{peng2017visda, peng2018visda}, object detection~\cite{hou2020multiview, gaidon2016virtual, cabon2020vkitti2}, multi-object tracking~\cite{gaidon2016virtual, cabon2020vkitti2, fabbri2018learning}, semantic and instance segmentation~\cite{sankaranarayanan2018learning, gaidon2016virtual, cabon2020vkitti2}, pose estimation~\cite{fabbri2018learning, doersch2019sim2real} and navigation~\cite{kolve2017ai2}. Commonly used simulation platforms include Unity and Unreal. In this area, domain adaptation is mostly used. For example, Bak \emph{et al.}~\cite{bak2018domain} use the cycle generative adversarial network to covert synthetic images into the real-world style. 
In comparison, there are much fewer works that do not need domain adaptation to get good performance in this area.
In our work, we investigate the possibility of using synthetic data to learn association modules in MOT.


\textbf{Domain gap beyond appearance.} While domain gap caused by image appearance is most studied, there are some works studying other factors that lead to distribution differences between domains.  
Recently, Meta-sim~\cite{kar2019meta} optimizes the probability grammar for scene content generation. 
Yao \textit{et al.}~\cite{yao2019simulating} study the content-level domain gap in the vehicle re-identification task and show that the feasibility to reduce such gap by editing synthetic data. This paper will identify and discuss factors beyond appearance (\emph{i.e.,} motion factors) that influence association learning in MOT.

\section{MOTX Engine}
MOTX is a 3D rendering engine that receives a set of controllable factors related to objects, cameras and others as inputs, and outputs a 2D video together with ground truth annotations (Fig.~\ref{fig:motx_data}). 
We build MOTX based on the Unity~\cite{juliani2018unity} game engine. Section~\ref{sec:cfactor} introduces controllable factors. Section~\ref{sec:motx_annotation} describes annotation acquirement.

\subsection{Controllable Factors} 
\label{sec:cfactor}

\textbf{Object-related factors.} Currently MOTX focuses on tracking pedestrian. We collect 1,200 pedestrian 3D models with distinct appearances from the PersonX engine~\cite{sun2019dissecting}. Controllable factors include pedestrian density, speed and action. Density refers to the number of pedestrians inside the viewing frustum.
Each pedestrian takes an action from $\{\texttt{walk}, \texttt{run}\}$ with a random speed drawn from a given speed distribution.  Walking routes are randomly generated. 

\textbf{Camera-related factors.} 
The viewing pose, spatial location, running path and speed of the camera can be flexibly adjusted.
In this paper, we mainly evaluate two commonly encountered camera views, the surveillance view (static camera, overlooking view), and the vehicle-mounted view (moving camera, near-horizontal view). 


\textbf{Others factors.}  MOTX supports changing other visual factors that can influence the final rendering, including scenes, resolution, lighting (light direction, light intensity and light color, \emph{etc}). If not specified, all videos are recorded at the resolution of 1024$\times$768. 


\subsection{Annotation Acquisition}
\label{sec:motx_annotation}
\textbf{Bounding box annotation.} We transform 3D locations of person models in the scene into 2D locations in the camera view. By calculating the top, bottom, left and right vertices of persons, we can obtain accurate bounding boxes for the holistic body. For occluded or partial visible persons, the engine can tell the occlusion relations and we accordingly annotate the bounding boxes of visible parts as well. 

\textbf{Identity annotation.} Identity labels are directly given by the engine.
This avoids the re-labeling problem when a person leaves and then re-enter the field of view, which is a common annotation mistake in real datasets.

\section{Association Knowledge}
A multiple object tracker is usually composed of a detector, an appearance model and an association model. In this work, we argue that it is possible to learn the association model with synthetic videos generated by the MOTX engine, while the learned association knowledge is applicable to real-world data without domain adaptation. As preliminary, we give a definition on the association knowledge and briefly review how existing methods learns it.

\subsection{Definition of Association Knowledge}

Given a set of detected bounding boxes $\mathcal{D}_{t}$ and tracked objects $\mathcal{O}_{t}$ at frame $t$, the assignment between the $i$-th bounding box $d_{i}$ in $\mathcal{D}_{t}$ and the $j$-th object $o_{j}$ in $\mathcal{O}_{t}$ is noted as $a_{ij}$, where $a_{ij} \in \{0,1\}$. $a_{ij} = 1$ denotes that $d_{i}$ is associated with $o_{j}$. Otherwise, $d_{i}$ belongs to other tracklets. The association module in an MOT systm usually aims to optimize the assignment matrix $\mathcal{A}$ at frame $t$:
\begin{equation}
\begin{split}
\footnotesize
\mathcal{A}_{t}^{*} = \text{argmax}_{\mathcal{A}_{t}}\sum_{i=1}^{\left | \mathcal{D}_{t} \right |}\sum_{j=1}^{\left | \mathcal{O}_{t} \right |}a_{ij}s_{ij},~~~~~~~~~~~~~ \\
s.t. \mbox{ } \mathcal{A}_{t} \in {\{0,1\}}^{\left | \mathcal{D}_{t} \right | \times \left | \mathcal{O}_{t} \right |}; ~\sum_{i=1}^{\left | \mathcal{D}_{t} \right |}a_{ij}\leqslant 1 ~ ; ~ \sum_{j=1}^{\left | \mathcal{O}_{t} \right |}a_{ij}\leqslant 1,
\end{split}
\label{assignment}
\end{equation}
where $a_{ij}$ is the entry of in $\mathcal{A}_{t}$ and $s_{ij}$ is the association score between $d_i$ with $o_j$. If $\sum_{i=1}^{\left | \mathcal{D}_{t} \right |}a_{i,j} = 0$, none of the bounding boxes in $\mathcal{D}_{t}$ should be connected to $o_j$. Similarly, $\sum_{j=1}^{\left | \mathcal{O}_{t} \right |}a_{ij} = 0$ indicates that bounding box $d_i$ does not belong to any objects in $O_t$. In this case, $d_i$ can be a new object, or the object ID that $d_i$ belongs to is missing in currently tracked objects. 

\begin{figure}[t]
	\centering  
	\includegraphics[width=0.9\linewidth]{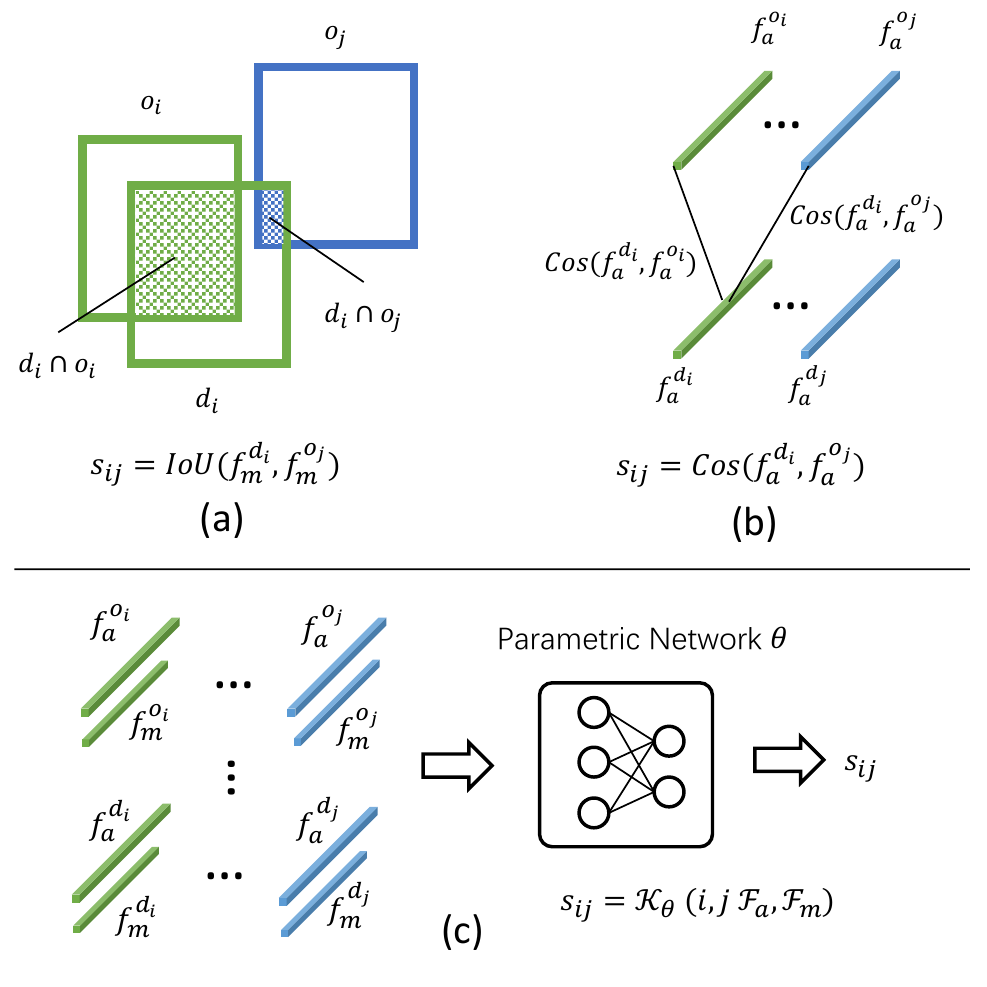}
	\caption{Illustration of different types of association knowledge. \textbf{(a)} and \textbf{(b)}: human-designed association knowledge. They compute the association score $s_{ij}$ using the IoU score of bounding boxes and the cosine similarity between re-ID features, respectively. \textbf{(c)}: uses the parametric network $\theta$ to predict the association score.
	}
 	\vspace{-1mm}
	\label{fig:association_knowledge}
\end{figure}

We define association knowledge  $\mathcal{K}$ as a metric function that takes appearance features, motion features or both of them as input and output the similarity score, 
\begin{equation}
s_{ij}=\mathcal{K}(i, j, \mathcal{F}_a, \mathcal{F}_m), 
\label{Eq_defineK}
\end{equation}
where $\mathcal{F}_a = \{f_a(d_1),...f_a(d_{\vert\mathcal{D}_{t}\vert}), f_a(o_1),...f_a(o_{\vert\mathcal{O}_{t}\vert})\}$ is the joint set of appearance features from both detections and existing tracklets, and $\mathcal{F}_m$ is similar but contains motion features. In practice, appearance features are widely represented by the Re-ID features  while motion features usually contains geometric information such as locations and sizes of bounding boxes~\cite{braso2020learning,li2020graph}. 


In early literature, the association knowledge $\mathcal{K}$ is commonly modeled with human-designed policies. 
For instance, a simple policy is to only consider motion cues $\mathcal{F}_m$, ignoring appearance cues $\mathcal{F}_a$.
Specifically, bounding boxes belonging to the same object ID in two adjacent frames should be closer than those belonging to different object IDs. Based on this observation, we can use the the Intersection over Unions (IoU) of the bounding boxes as the association score (Fig.~\ref{fig:association_knowledge} (a)).
Another simple yet effective human-designed policy is to use  the cosine similarity between the Re-ID features as  the association score. Similarly, the cosine similarity belonging to the same ID have larger value than that computed from Re-ID features extracted from different identities (Fig.~\ref{fig:association_knowledge} (b)). 

Human-designed policies are sub-optimal as it is difficult for them to take full advantage of both appearance and motion cues.
Beyond human-designed policies, more recent arts~\cite{braso2020learning, li2020graph,xu2020train, papakis2020gcnnmatch} attempt to learn association knowledge directly from data with a parametric model, \ie, $s_{ij}=\mathcal{K}_{\boldsymbol{\theta}}(i,j,\mathcal{F}_a, \mathcal{F}_m)$.
As illustrated in Fig.~\ref{fig:association_knowledge} (c), 
both $\mathcal{F}_a$ and $\mathcal{F}_m$ are taken as input by the association model, and the model learns it parameter $\boldsymbol{\theta}$ by applying Stochastic Gradient Descent (SGD) on a labelled dataset. During inference, $\mathcal{K}_{\boldsymbol{\theta}}$ output predictions with a single forward pass. A most prevalent choice of the parametric model is the Graph Neural Network (GNN)~\cite{scarselli2008graph}.
In the next section, we show by empirical experiments that it is possible to learn association knowledge from synthetic data.

\begin{figure*}[t]
	\centering  
	\includegraphics[width=\linewidth]{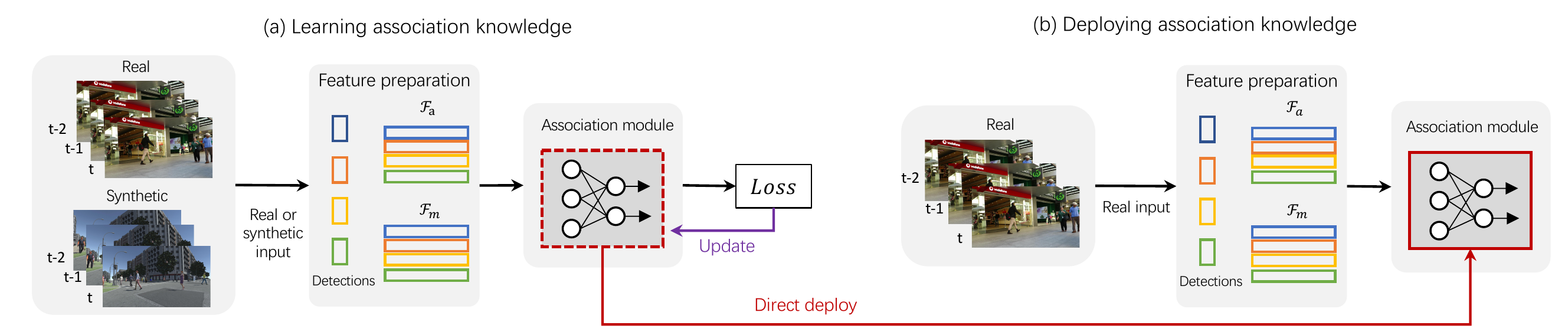}
	\caption{Comparison pipeline. (\textbf{a}) For each association module, we train or tune two association modules with synthetic data and real data respectively.  (\textbf{b}) We then directly deploy the two association modules on a real-world test set and compare their performance. $\mathcal{F}_a$ and $\mathcal{F}_m$ are appearance features and motion features, respectively. 
	}
 	\vspace{-1mm}
	\label{fig:learning_pipeline}
\end{figure*}





\section{Experiment Setup}
\textbf{Comparison pipeline.} This paper aims to compare synthetic data and real data on their effectiveness when they are used to learn association knowledge. 
The experimental setup is briefly illustrated in Fig.~\ref{fig:learning_pipeline}. The association algorithm is trained on real data and synthetic data respectively, and finally, test both of them on the real-world set. Note that when an algorithm involves multiple models to be trained, we only train the association-related model and keep other models fixed. For instance, if an algorithm employs a Re-ID model, we fix the Re-ID model.  During inference, we do not perform domain adaptation.

\textbf{Benchmark methods.} For a comprehensive comparison, 
we select several typical association methods include both parametric association models and human-designed association policy. We pay more attention to parametric models as they show superior performance. Details are described as follows: 
\textbf{MPNTracker}~\cite{braso2020learning} formulates MOT with the classical network flow. A type of GNNs named Message Passing Networks (MPNs) are proposed to predict linkages based on the graph built with appearance features and motion cues. 
\noindent \textbf{DeepMOT}~\cite{xu2020train} proposes a Deep Hungarian Net (DHN) as the association module to approximates the Hungarian matching algorithm. 
\noindent \textbf{GNMOT}~\cite{li2020graph} builds the appearance graph and the motion graph for two conjunctive frames. Then two graph networks compute the similarities between nodes to achieve association. 
\noindent \textbf{SORT}~\cite{bewley2016simple} is a human crafted association policy. It only employs the motion cues, observations are associated with tracklets in a hierarchical manner by comparing IoU distances. We mainly tune the key hyper-parameter, IoU threshold in the training set and use it in the test set.

\textbf{Evaluation metric.} For evaluation, we employ the widely used CLEAR~\cite{bernardin2008evaluating} metrics. Main metrics including MOTA (MOT Accuracy), IDF1 (ID F1-Measure), IDSwR (Identity Switch Rate), MT (Mostly Tracked Target Percentage), and ML (Mostly Lost Target Percentage). Among them, IDF1 and IDs are the most relevant ones to evaluate association accuracy.

\section{Results and Analysis}

\begin{table}[t]
			\centering

			\setlength{\tabcolsep}{1mm}{
			\footnotesize
				\begin{tabular}{|l|l|ccccc|}
					\hline	
					Test & Train & MOTA$\uparrow$ & IDF1$\uparrow$ & IDSwR$\downarrow$ & MT$\uparrow$ & ML$\downarrow$ \\
					\hline
                    \hline
					\multicolumn{7}{|c|}{DeepMOT} \\
					\hline
					
					\multirow{2}{*}{MOT16} & MOT16-train & 54.8 & 53.4 & 11.4 & 19.1 & 37.0 \\
                    &MOTX-S & \underline{54.4} & \underline{53.2} & \underline{12.1} & \textbf{19.2} & \underline{37.2} \\
                    \hline
					\multirow{2}{*}{MOT17} & MOT17-train & 53.7 & 53.8 & 34.7 & 19.4 & 36.6  \\
                    &MOTX-S & \underline{53.4} & \underline{52.9} & 36.4 & \textbf{19.7} & \textbf{36.6} \\
                    \hline
                    \hline
					\multicolumn{7}{|c|}{GNMOT} \\
					\hline
					\multirow{2}{*}{MOT16} & MOT16-train & 58.4 & 54.8 & 23.3 & 27.3 & 23.2  \\
                    &MOTX-S & \textbf{58.4} & \underline{54.5} & \underline{23.6} & \textbf{27.3} & \underline{23.3}  \\
                    \hline
					\multirow{2}{*}{MOT17} & MOT17-train & 56.9 & 53.9 & 72.2 & 25.9 & 25.6  \\
                    &MOTX-S & \underline{56.8} & \underline{53.6} & \underline{73.0} & \textbf{26.1} & \underline{25.7} \\
                    \hline
                    \hline
					\multicolumn{7}{|c|}{MPNTracker} \\
					\hline
					\multirow{2}{*}{MOT15} & MOT15-train & 51.5 & 58.6 & 5.8 & 31.2 & 25.9  \\
                    &MOTX-S & \underline{51.3} & \textbf{59.1} & \textbf{5.8} & \textbf{34.3} & \textbf{25.2} \\
                    \hline
					\multirow{2}{*}{MOT17} & MOT17-train & 58.8 & 61.7 & 6.0 & 28.3 & 33.5  \\
                    &MOTX-S & \underline{58.4} & \underline{61.0} & \underline{6.1} & \underline{28.1} & \underline{33.8}  \\
                    \hline
                    \hline
					\multicolumn{7}{|c|}{SORT} \\
					\hline
					\multirow{2}{*}{MOT15} & MOT15-train & 42.6 & 50.8 & 7.27 & 11.2 & 37.6    \\
                    &MOTX-S & \textbf{42.6} & \underline{50.4}  & \underline{7.25} & \textbf{11.2} & \textbf{37.6} \\
                    \hline
				\end{tabular}
			}
			
			\caption{Comparing synthetic data (MOTX-S) and real data in association knowledge learning on real-world test sets. \textbf{Bold} numbers denote association knowledge learned from synthetic data are superior or equal to that learned from real data, while  and \underline{underlined} ones mean the performance gap is less than 1.0.} 
			\label{tab:benchmark_testing}
			\vspace{-2mm}
\end{table}

\subsection{Evaluation on Benchmark Datasets}
\label{Sec:main_result}

In this section, we show association knowledge learned from synthetic data works well on real-world test sets. Specifically, we use the \emph{test} set of MOT-15/16/17~\cite{leal2015motchallenge, milan2016mot16}. For real data training, we use the corresponding \emph{train} split of the target set, \eg, train on MOT16 \emph{train} and test on MOT16 \emph{test}.  For synthetic data training, we build a synthetic training set, use this single set for training, and evaluate on all test sets. We name the synthetic dataset MOTX-S. MOTX-S is synthesized using the MOTX engine, consisting of 22 videos in total. Videos are generated by roughly simulating the scene dynamics (camera moving, camera view, person density, person velocity, \etc) of videos in MOT15-17 dataset. As shown in Section~\ref{sec:cfactor}, the resulting synthetic videos yield consistently good results even when parameters of some scene dynamics vary in a relatively large range.

\textbf{Association Knowledge from the synthetic world is competitive.} For the generality of the results, we test with multiple different association algorithms. Results are reported in Table~\ref{tab:benchmark_testing}. The major observation is that each association method trained on synthetic data can have similar performance to that trained on the real-world training data in terms of all metrics. Note that, when training MPNTracker, MOTX-S shows its advantage over MOT15.
Specifically, MOTX-S makes improvements IDF1, MT, and ML with 0.5$\%$, 3.1$\%$, and 0.7$\%$, respectively. It suggests that the association scenarios in MOTX give better supervision on association knowledge learning than MOT15.
For all comparisons, we do not observe a noticeable performance drop when trained on MOTX-S. In most cases, the performance gap between MOTX and the real-world data is less than 1$\%$ over all evaluation indexes. 
The above observations suggest that the association knowledge learned from synthetic data can achieve similar performance compared with that trained on real-world data. On the other hand, such competitiveness of synthetic data can not be seen in ``Appearance-centered tasks'' if the deep system is only learned from the synthetic data. Because of the superior performance and run-time efficiency of MPNTracker, experiments in Section~\ref{sec:cfactor} is conducted on it.

\begin{table}[t]
			\centering
			\setlength{\tabcolsep}{1mm}{
			\footnotesize
				\begin{tabular}{|l|l|ccc|}
					\hline	
					Test& Train & MOTA$\uparrow$ & IDF1$\uparrow$ & $\#$ IDS$\downarrow$ \\
					\hline
					\hline
					\multirow{4}{*}{MOT15-test} & MOT15   & \textbf{51.5} & 58.6 & \textbf{375} \\
                    & MOT17 & 50.9 & 58.8 & 381  \\
                    & MOT15+MOT17 & 51.3 & 58.9 & 382 \\
                    & MOTX-S & 51.3 & \textbf{59.1} & 377 \\
                    \cline{1-5}
                    \multirow{4}{*}{MOT17-test} & MOT17 &  \textbf{58.8} & \textbf{61.7} & \textbf{1185} \\
                    & MOT15 & 57.9 & 58.7 & 1481 \\
                    & MOT17+MOT15 & 58.3 & 60.8 & 1267 \\
                    & MOTX-S & 58.4 & 61.0 & 1214 \\
                    \hline
                    
				\end{tabular}
			}
			\caption{Cross domain evaluation. \textbf{Bold} numbers denote the best results.} 
			\label{tab:cross_domains}
			\vspace{-3mm}
\end{table}

\begin{table}[t]
			\centering
			\setlength{\tabcolsep}{1mm}{
			\footnotesize

				\begin{tabular}{|l|l|ccc|}
					\hline	
					Test & Train & MOTA$\uparrow$ & IDF1$\uparrow$ & $\#$ IDS$\downarrow$ \\
					\hline
					\hline
                    \multirow{2}{*}{MOT17-train} &  MOTX-S & \textbf{64.1} & \textbf{68.9} & \textbf{551}  \\
                     & MOTX-S + SPGAN & 63.7 & 68.2 & 604\\
                    \cline{1-5}
                    \multirow{2}{*}{MOT15-train} & MOTX-S & \textbf{53.1} & \textbf{67.9} & \textbf{78}  \\
                    & MOTX-S + SPGAN & 52.5 & 66.6 & \textbf{78}  \\
                    \hline
                    
				\end{tabular}
			}
			\caption{Impact of appearance domain adaptation for pedestrians. Best results are marked as \textbf{Bold}.} 
			\label{tab:appearance_DA}
			\vspace{-3mm}
\end{table}

\textbf{Association domain gap exists.}
We train MPNTracker on the training set on MOT15, MOT17, and their combination, respectively. Testing results on MOT17 test set are shown in Table~\ref{tab:cross_domains}. Both MOT15 and the combined set are worse than using MOT17 only. Specifically, MOT15 get 3$\%$ lower IDF1 and about 25$\%$ higher ID switches. 

 A similar degeneration trend can also be found when deploying the association knowledge from MOT17 into the MOT15 domain. This suggests that there is a domain gap between association scenarios in MOT15 and MOT17.

\textbf{Appearance domain adaptation is not necessary.}
We attempt to reduce the appearance domain gap between synthetic data and real-world data by converting the appearance of detections in MOTX-S into the real-world style by using a generative network SPGAN~\cite{deng2018image}. SPGAN is trained on data provided  by VisDA2020\footnote{http://ai.bu.edu/visda-2020/}, which has both Unity-based synthetic persons and real-world persons. Results in Table.~\ref{tab:appearance_DA} show that MOTX-S is still competitive without domain adaptation on appearance.

\begin{figure}[t]
	\centering  
	\includegraphics[width=\linewidth]{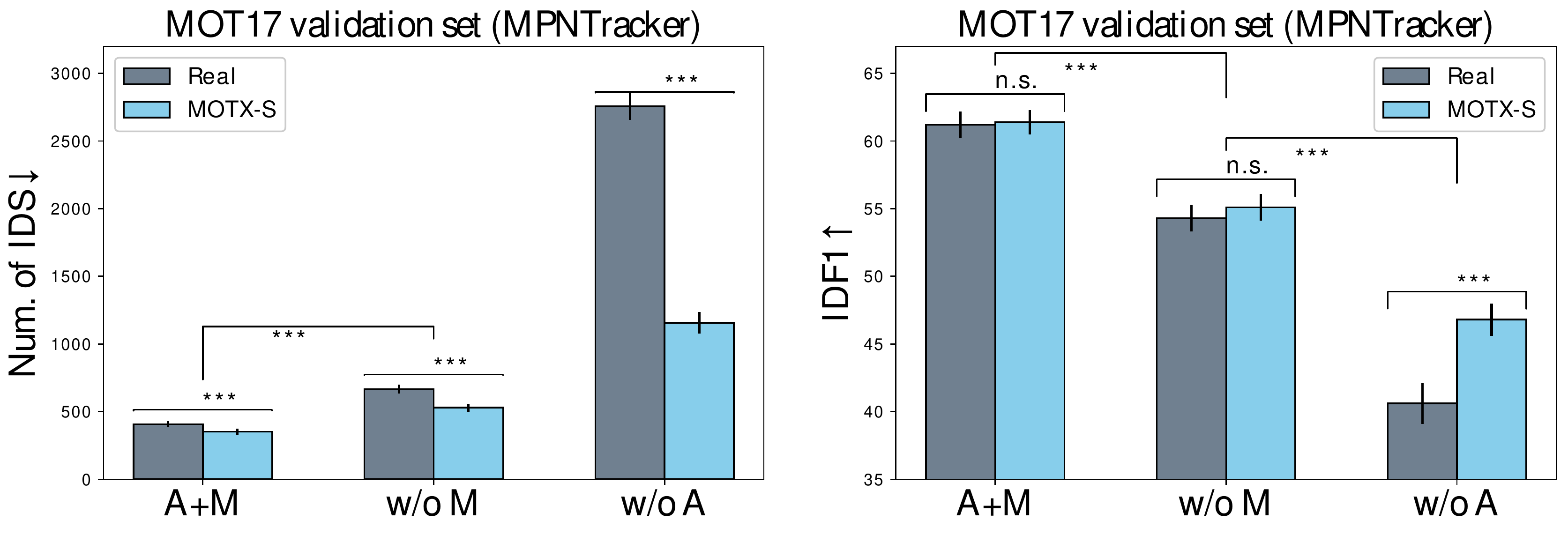}
	\caption{Ablation study on appearance features $\&$ motion features. \textbf{A+M}: with both appearance and motion features; \textbf{w/o A}: without appearance features;   \textbf{w/o M}: without motion features. ``n.s.'' means the difference is {not statistically significant} ($i.e., p$-value $>$ 0.05).  $**$ and $***$ mean {statistically very significant} ($i.e., 0.001 < p$-value $< 0.01$) and {statistically extremely significant} ($i.e., p$-value $< 0.001$), \emph{resp}.
	}
 	\vspace{-1mm}
	\label{fig:ablation_am}
\end{figure}

\subsection{Ablation Study on $\mathcal{F}_a$ And $\mathcal{F}_m$}
It is worthwhile to investigate why the competitive results in Table~\ref{tab:benchmark_testing} can be achieved by only using synthetic data with a considerable domain gap on image-style. We conduct the ablation study on the input of the association model. Specifically, we eliminate the effect of appearance features $\mathcal{F}_a$ or motion features $\mathcal{F}_m$ in Equation~\ref{Eq_defineK} by replacing them with dummy vectors $\boldsymbol{1} = (1,\dots,1)^{T}$. 
Videos \{2, 10, 13\} in MOT17 are split as the validation set and the rest videos make up the training set.
We repeat each training on MPNTracker for 5 times and report their means. We also perform hypothesis testing to validate the statistical significance of the results. Results are shown in Fig.~\ref{fig:ablation_am}. 

\textbf{Effectiveness of appearance features and motion features.}
The tracking performance degenerates when we eliminate either appearance features or motion features. It shows that both appearance features and motion features contribute to association knowledge learning. When training on both appearance $\&$ motion features, MOTX-S achieves similar performance on MOT17 validation set. This is consistent with the conclusion we get in Section~\ref{Sec:main_result}.

\textbf{Synthetic \emph{vs.} real on motion features.} When only motion features are used (w/o A), MOTX-S shows a considerable advantage over real data. In detail, the ID switch for MOTX-S is only half of that for real data. IDF1 score also leads by over 6$\%$. Such performance gap is not observed in  experiments ``A+M'' and ``w/o M''.  This phenomenon suggests that motion scenarios generated with MOTX engine can simulate the real-world association scenarios well.

\textbf{Synthetic \emph{vs.} real on appearance features.}
Intuitively, it is highly possible that the domain gap of the appearance feature harms association learning. This is because appearance models are trained on real-world Re-ID datasets, but in training association models, they are used to extract features of synthetic person images. Moreover, the final test set consists of real-world videos. However, we do not observe the expected performance drop due to the appearance domain gap.
 See results ``w/o M'' in Fig.~\ref{fig:ablation_am}, with appearance cues only, trained on real data and synthetic data performs almost equally with similar IDs and IDF1. This suggests a somehow surprising finding: \emph{Appearance domain gap hardly harms the learning of association knowledge}.
 

\begin{table}[t]
			\centering
			\setlength{\tabcolsep}{1mm}{
			\scriptsize
			\resizebox{0.475\textwidth}{!}{
				\begin{tabular}{|c|c|c|c|} 
                    \hline
                    \# &  & Notation & Description \\ 
                    \hline
                    \multirow{4}{*}{1} &\multirow{2}{*}{\rotatebox[origin=c]{90}{train}}& S-Cam-H& camera view: High (surveillance view) \\ 
                    \cline{3-4} 
                    & &S-Cam-L& camera view: Low (vehicle view) \\ 
                    \cline{2-4}
                    &\multirow{2}{*}{\rotatebox[origin=c]{90}{test}}& R-Cam-H & video \#04 in MOT17  \\
                    \cline{3-4}
                    & & R-Cam-L & video \#02, \#09 in MOT17 \\ 
                    \cline{1-4}
                    \multirow{4}{*}{2}&\multirow{2}{*}{\rotatebox[origin=c]{90}{train}}& S-Cam-S& camera state: Static \\
                    \cline{3-4}
                    & &S-Cam-M& camera state: Moving \\
                    \cline{2-4}
                    &\multirow{2}{*}{\rotatebox[origin=c]{90}{test}}& R-Cam-S & video \#02, \#04, \#09 in MOT17 \\ 
                    \cline{3-4}
                    && R-Cam-M & video \#10, \#11, \#13, in MOT17 \\ 
                    \cline{1-4}
                    \multirow{4}{*}{3}&\multirow{2}{*}{\rotatebox[origin=c]{90}{train}} & \multirow{2}{*}{S-Speed-$n$} & \multirow{2}{*}{pedestrian speed: $n$ m/s, $n \in \{ 1,2,4,6\}$}\\ 
                    & & & \\
                    \cline{2-4} 
                    & \multirow{2}{*}{\rotatebox[origin=c]{90}{test}}& R-Speed-H &  KITTI-17, KITTI-13, PETS09-S2L1 in MOT15  \\ 
                    \cline{3-4}
                    & &R-Speed-L &  Venice-2, ADL-Rundle-8, ADL-Rundle-6 in MOT15 \\ 
                    \cline{1-4}
                    \multirow{5}{*}{4}&\multirow{2}{*}{\rotatebox[origin=c]{90}{train}} & \multirow{2}{*}{S-Density-$n$} & \multirow{2}{*}{the number of persons in a frame, $n \in \{10,20,40\}$}\\
                    & & & \\
                    \cline{2-4}
                    & \multirow{3}{*}{\rotatebox[origin=c]{90}{test}} &\multirow{2}{*}{R-Density-L} & PETS09-S2L1, TUD-Stadtmitte, TUD-Stadtmitt,\\ 
                    & &&TUD-Campus, KITTI-17, KITTI-13 in MOT15 \\ 
                    \cline{3-4}
                    & &R-Density-H & video \#02, \#04 in MOT17 \\ 
                    \hline
                \end{tabular}
			}
			}
			\caption{Notations for 4 groups of data to study motion factors. The prefix ``S'' and ``R'' represent synthetic data and real data, respectively. The suffix ``H'', ``L'', ``S'', ``M'' stands for high, low, static and moving.}
			\label{tab:data_notations}
\end{table}

\begin{figure*}[t]
	\centering  
	\includegraphics[width=\linewidth]{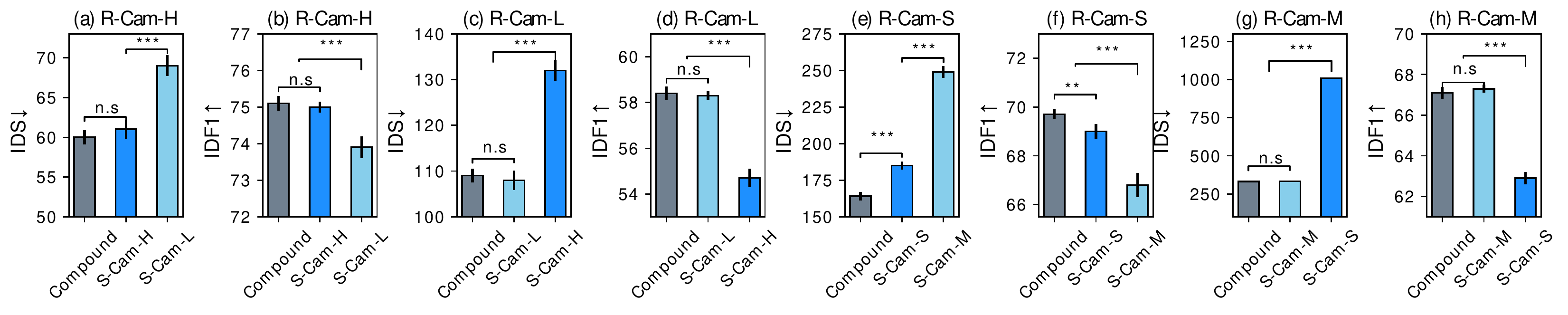}
	\caption{Impact of motion factors related to the camera. (a)-(d) illustrate the learning system reacts to the change of camera view. Similarly, (e)-(h) gives the result of changing the camera's moving state. All experiments are tested on selected real-world videos. Notations ``n.s.'', **, and *** have the same meaning as those in Fig.~\ref{fig:ablation_am}.
	}
 	\vspace{-1mm}
	\label{fig:factor_camera}
\end{figure*}

\textbf{Discussion.} The above insightful findings suggest that it is not necessary to perform additional appearance adaptation techniques when we deploy the learned association knowledge in the real-world test sets.
Also, we observe that our synthetic data show stronger competitiveness in Fig.~\ref{fig:ablation_am} than that in Table~\ref{tab:benchmark_testing}. The possible reason is that the training set and test set in MOTChallenge benchmark has overlap on association scenarios, \emph{i.e,} videos in the test set are collected at the same location as the training set where the camera-related and pedestrian-related factors are very close.

\subsection{Investigation on Controllable Factors}
\begin{figure}[t]
	\centering  
	\includegraphics[width=\linewidth]{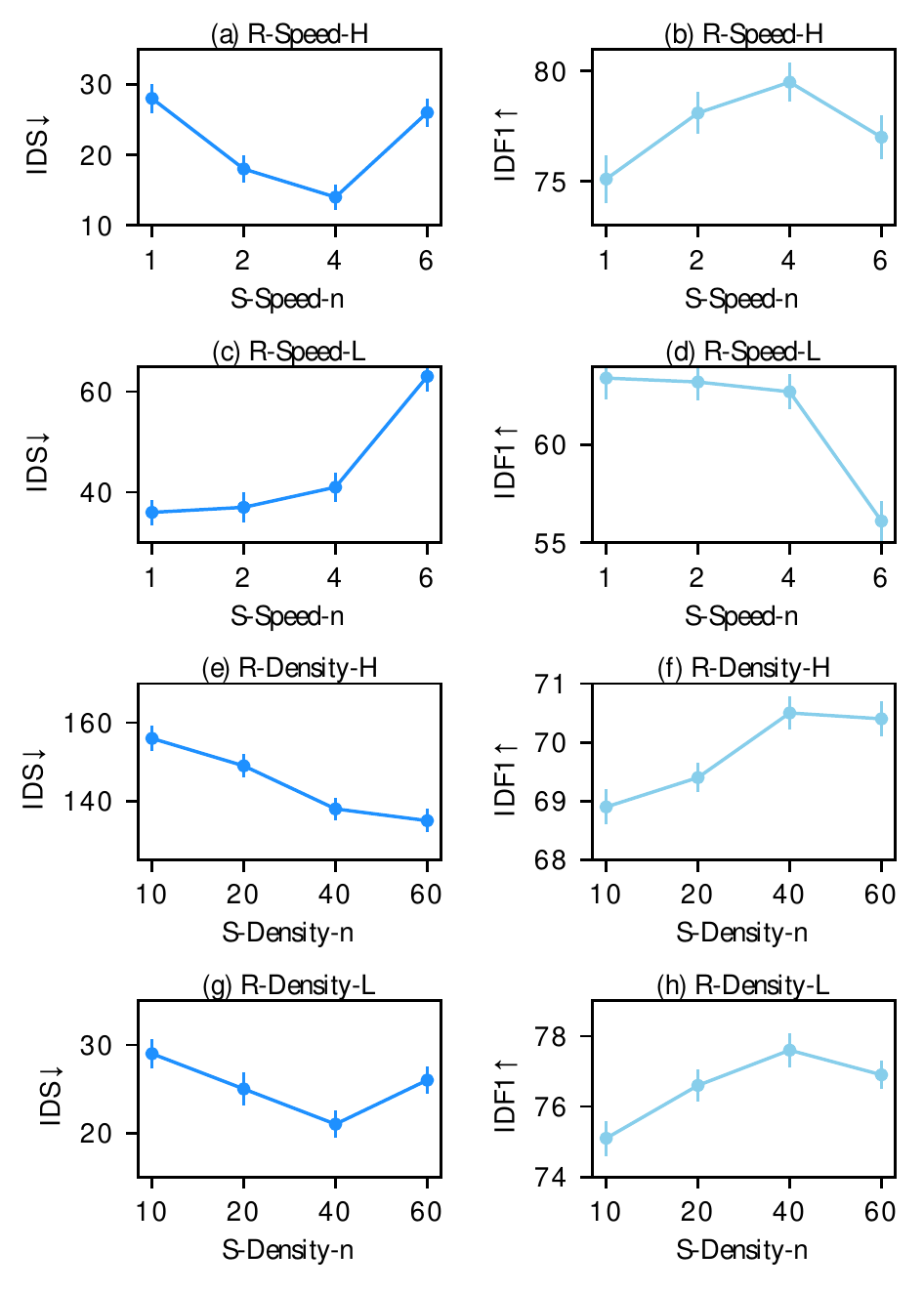}
	\vspace{-8mm}
	\caption{Impact of pedestrian-related motion factors. (a)-(d) illustrates the learning system reacts to the pedestrian speed. Similarly, (e)-(h) gives the result of changing the pedestrian density. All testing videos are real-world videos.
	}
 	\vspace{-1mm}
	\label{fig:factor_pedestrian}
\end{figure}

Another major advantage of synthetic data is that we can control multiple factors in generating videos. Therefore it is possible to conduct a thorough investigation on how these controllable factors impact an association algorithm with the help of synthetic data. In this section, we mainly study the influence of four factors,\ie, camera view, camera moving state, pedestrian speed, and pedestrian density. For each factor, we design a group of contrast experiments using different training sets and test sets. A summary of the used datasets is illustrated in Tab.~\ref{tab:data_notations}. A principle of these experiments is that we train an identical association model (here we use MPNTracker) with different customized \emph{synthetic} data (\eg, camera view high \vs low), and test on different \emph{real} data (both camera view high \vs low). Experimental results are shown in Fig.~\ref{fig:factor_camera} and Fig.~\ref{fig:factor_pedestrian}.

\textbf{Dataset notation.} For clarification, datasets are notated in format \texttt{prefix-middle-suffix}. The prefix can be ``S'' and ``R'', representing synthetic data or real data. The middle word is the controllable factor to be studied, \eg, ``Cam'' indicates camera. The suffix is the value of the controllable factor. For instance, ``S-Cam-H'' represents this dataset consists of synthetic videos with high camera view.


\textbf{Camera view.} The association models are trained on S-Cam-H, S-Cam-L and their compound version \{S-Cam-H, S-Cam-L\}, respectively. Then the trained association models are tested on real-world videos with high camera view (R-Cam-H) or low camera view (R-Cam-L). According to Fig.~\ref{fig:factor_camera} (a)-(d), a major observation is that association knowledge learning is sensitive to camera view.
Specifically, when testing on R-Cam-H, the association model trained on S-Cam-H can achieve a close ID switch rate and IDF1 score compared with the model trained on the compound data. HoweverOtherwise, the accuracy of only using S-Cam-L decreases noticeably in this case (a)-(b). We observe a similar trend when testing on videos with low camera view in (c)-(d). This suggests that the knowledge learned from high camera view can not be deployed in the low camera view test environment successfully, and vice versa.
In another word, there is an obvious association domain gap between high camera view scenarios and low camera view scenarios. 

\textbf{Camera moving state.} We learn association knowledge from static cameras (S-Cam-S) and moving cameras (S-Cam-M) and their combination (\{S-Cam-S, S-Cam-M\}). The real-world videos in two test sets are selected from MOTchallenge. One constrains videos as static cameras (R-Cam-S) and cameras in another are moving (R-Cam-M). Results are shown in Fig.~\ref{fig:factor_camera} (e)-(h). The same trend with camera view is that the camera moving state can also bias the association knowledge learning. For instance, (e)-(f) shows that S-Cam-S has the advantage in testing videos with static cameras. Similarly, the model trained on S-Cam-M obtains better results on R-Cam-M than that trained on S-Cam-S ( (g)-(h)). So, we can conclude that the moving state of cameras can cause the domain gap in association scenarios. 
Also, we observe an obvious performance increase on IDS and IDF1 if we combine S-Cam-M with S-Cam-S in (e-f). However, such a trend is not observed in (g)-(h) where the test set is R-Cam-M. 
This insightful discovery implies the association knowledge learned from moving cameras have stronger compatibility than that learned from static cameras. 

\textbf{Pedestrian speed.} Association models are trained on S-Speed-$n$, $n \in$\{1, 2, 4, 6\}, which means pedestrian speed is $n$ m/s. Test sets are R-Speed-L and R-Speed-H. In detail, the frame rate of videos in R-Speed-H ranges from 7 fps to 10 fps. It means that the moving speed of the same identity between two conjunctive frames is almost 3-4 times as that in R-Speed-L where the video frame rate is around 30 fps.
According to Fig.~\ref{fig:factor_pedestrian} (a)-(d), we have two observations. The major one is that if the pedestrian speed in the training set mismatches speeds in the test environment, the performance decreases. 
when testing on R-Speed-H, the number of IDs doubled (14.2 $\rightarrow$ 28.6) if the pedestrian speed is changed from 4 m/s to 1 1m/s.  The IDF1 score also degenerates obviously (79.5$\%$ $\rightarrow$ 75.1$\%$). A similar trend can be found when testing on R-Speed-L. Accelerating the pedestrians' speed to large values (1 m/s $\rightarrow$ 6m/s) can significantly increase the number of IDs and decrease IDF1.

\textbf{Pedestrian density.}
Fig.~\ref{fig:factor_pedestrian} (e)-(h) shows results for testing on real videos with different pedestrian densities. According to official statistics in MOTChallenge,  the average pedestrian density of all videos in our build R-Density-L is less than 10. However, for R-Density-L, the optimal density in the training set is 40 according to (g)-(h). It suggests the gap in pedestrian density does not cause the gap in tracking performance if the testing environment has low pedestrian density. For example, S-Density-60 is better than S-Density-10 when testing on R-Density-L. However, association knowledge gained from low-density videos is not very effective in high-density environments ((e)-(f)).

\textbf{Discussion.} Our synthetic dataset is manually configured in MOTX. We do so by setting the motion-related parameters to roughly match the real training videos. The above experiment also serves as a confirmation that this manual configuration process is stable. For example, when the pedestrian speed is set between 1-2 m$/$s, the IDS scores remain stable. The same observation also goes for other factors like pedestrian 
density. Therefore, in practice, we advise giving a possibly best manual estimation of the motion parameters of the environment to be tested. Relatively small errors can be well-tolerated, but large errors (\eg., the camera speed is estimated to be 1 m$/$s but is actually static) should be avoided.

\begin{table}[t]
			\centering
			\setlength{\tabcolsep}{1mm}{
			\small
				\begin{tabular}{|c|c|c|c|}
					\hline	
				    Dataset & S-Speed-H & MOT15-train & R-Speed-H \\
					\hline 
					IoU Threshold (0$\sim$1)  & 0.25 & 0.2 & 0.3\\

					\hline 
				\end{tabular}
			}
			\caption{Hyper-parameter tuning for SORT algorithm. We report the best hyper-parameter for the given dataset.}
			\label{tab:hyper_parameter}
\end{table}

\subsection{Tuning Human-designed Policy \wrt Scenes}
Our synthetic dataset can also benefit hyper-parameter search for the given scenes. We take the SORT~\cite{bewley2016simple} algorithm as an example. When R-Speed-H is the testing scenario, we synthesize a dataset according to the roughly estimated motion factors in R-Speed-H and search for the hyper-parameter IoU threshold. As shown in Table~\ref{tab:hyper_parameter}, comparing with using MOT15 for hyper-parameter search, IoU threshold searched from the synthetic data is closer to that searched from fully labeled R-Speed-H. 

\section{Conclusion}
This paper studies the role of synthetic data in Multi-Object Tracking. Crediting to the proposed MOTX engine, we make two contributions.
First, we show that association knowledge obtained from synthetic data can be directly deployed in the real-world environment without domain adaptation, even if the image-style discrepancy between synthetic data and real-world data exists. Second, with the help of MOTX engine, we thoroughly investigate how association knowledge reacts to changes of camera-related and pedestrian-related motion factors. Experimental results lead to intriguing finds giving new insights to understand the impact of data in association knowledge learning.

{\small
\bibliographystyle{ieee_fullname}
\bibliography{main}
}

\end{document}


\title{Supplementary Material}


\twocolumn[{%
\renewcommand\twocolumn[1][]{#1}%
\maketitle
\begin{center}
    \centering
    \captionsetup{type=figure}
    \includegraphics[width=\textwidth]{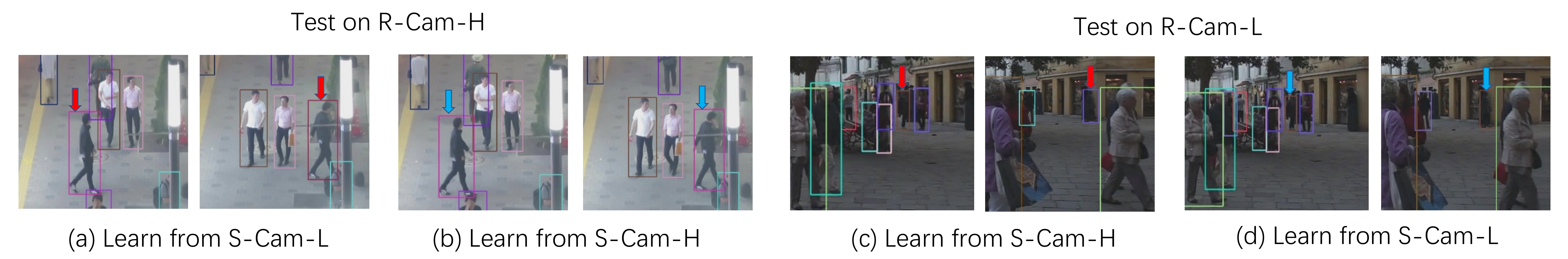}
    \captionof{figure}{Failure and success cases produced by association modules learned from synthetic training data with different camera views \{low view (
    \emph{i.e.,} vehicle view), high view (\emph{i.e.,} surveillance view)\}. \red{Red} arrows denote ID switches. \blue{Blue} arrows mean the ID is preserved cross video frames.}
    \label{fig:Vis_camera_V}
\end{center}%

\begin{center}
    \centering
    \captionsetup{type=figure}
    \includegraphics[width=\textwidth]{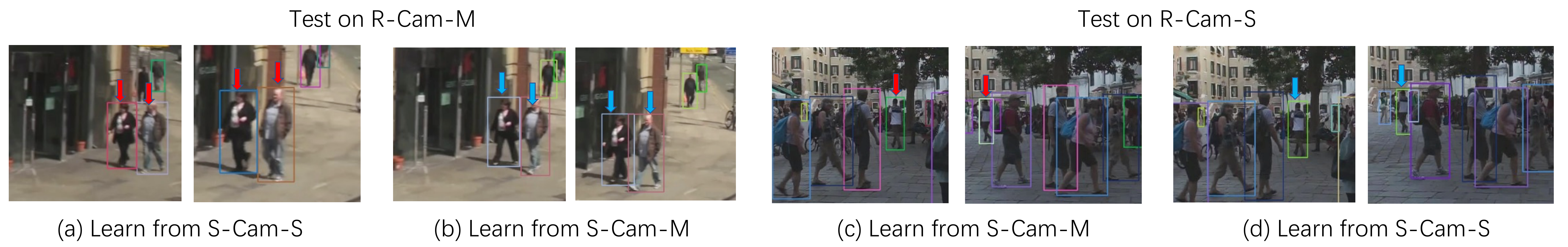}
    \captionof{figure}{Failure and success cases produced by association modules learned from synthetic training sets with different camera moving states \{static, moving\}. \red{Red} and \blue{Blue} arrows have the same meaning with Fig.~\ref{fig:Vis_camera_V}.}
    \label{fig:Vis_camera_M}
\end{center}%

\begin{center}
    \centering
    \captionsetup{type=figure}
    \includegraphics[width=\textwidth]{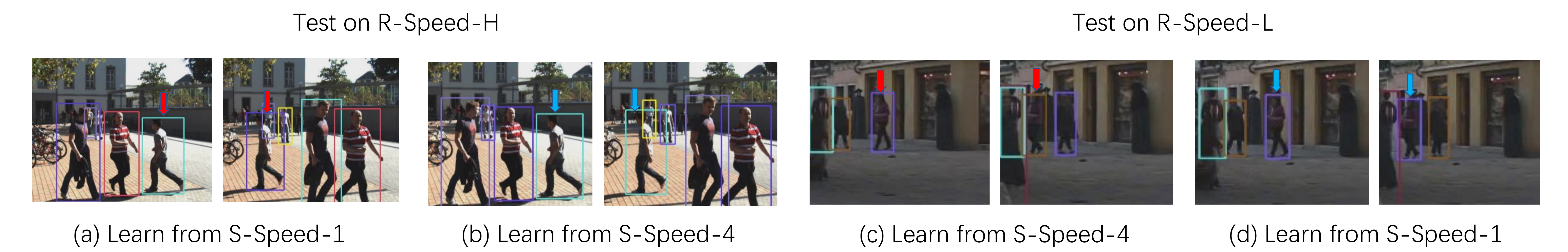}
    \captionof{figure}{Failure and success cases produced by association modules learned from training sets with different pedestrian speeds \{1 m/s, 4 m/s\}. \red{Red} and \blue{Blue} arrows have the same meaning with Fig.~\ref{fig:Vis_camera_V}.}
    \label{fig:Vis_pedestrian_S}
\end{center}%

\begin{center}
    \centering
    \captionsetup{type=figure}
    \includegraphics[width=\textwidth]{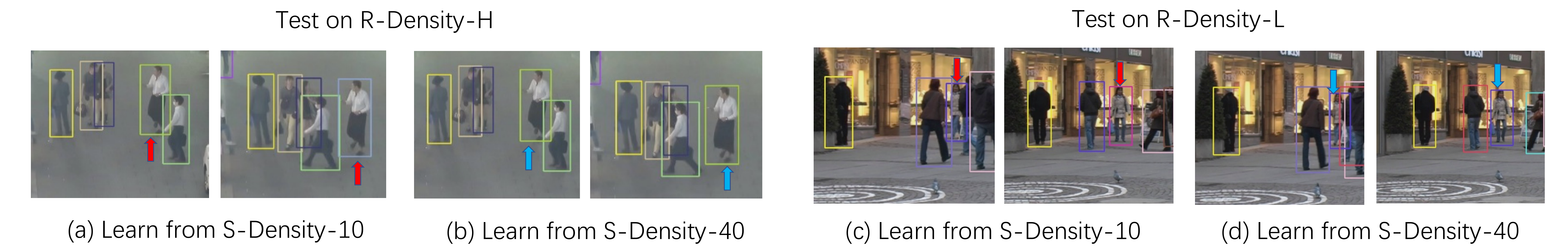}
    \captionof{figure}{Failure and success cases produced by association modules learned from synthetic training data with different pedestrian densities \{density = 10, density = 40\}. \red{Red} and \blue{Blue} arrows have the same meaning with Fig.~\ref{fig:Vis_camera_V}.}
    \label{fig:Vis_pedestrian_D}
\end{center}%

}]



\section{Examples of Synthetic Videos} 
We attach some examples of synthetic video clips in \texttt{.gif} format in the uploaded zip file. Their descriptions are provided in Table~\ref{tab:gif}.

\begin{table}[t]
			\centering
			\setlength{\tabcolsep}{3mm}{
				\begin{tabular}{|l|l|} 
                    \hline
                    GIF Name & Description  \\ 
                    \hline
                    Cam\_Low.gif & vehicle view \\
                    Cam\_High.gif & surveillance view \\
                    Cam\_Static.gif & camera is static. \\
                    Cam\_Moving.gif & camera is moving. \\
                    Ped\_Speed4.gif & pedestrian speed is 4 m/s.\\
                    Ped\_Density60.gif & pedestrian density is 60.\\
                    \hline
                \end{tabular}
			}
			\caption{Descriptions of synthetic video clips included in the supplementary materials.}
			\label{tab:gif}
\end{table}

\section{Visualization of Tracking results} 
We show some failure and success cases of our system in Fig.~\ref{fig:Vis_camera_V}, Fig.~\ref{fig:Vis_camera_M}, Fig.~\ref{fig:Vis_pedestrian_S}, and Fig.~\ref{fig:Vis_pedestrian_D}.
Different color indicates different identities.  All experiments use synthetic data for training and use real-world data for testing. Notations and corresponding descriptions for all training sets and test sets are provided in Section 6.3 in the main paper. 

Failure cases in Fig~\ref{fig:Vis_camera_V} and Fig.~\ref{fig:Vis_camera_M} suggest that view and moving state of cameras can bias the association knowledge learning. The gap in camera-related motion factors between training sets and test sets can give rise to ID switches. A similar phenomenon can be observed on pedestrian speed (Fig.~\ref{fig:Vis_pedestrian_S}) while association knowledge can be tolerant to the testing scenario with more sparse pedestrians than training sets.
Finds in the above figures are consistent with the results and analysis in Section 6.3 in the main paper.